\documentclass[a4paper,fleqn]{cas-dc}

\usepackage[authoryear,longnamesfirst]{natbib}

\def\tsc#1{\csdef{#1}{\textsc{\lowercase{#1}}\xspace}}
\tsc{WGM}
\tsc{QE}

\newtheorem{definition}{Definition}[section]

\usepackage{amsmath,amssymb,amsfonts,url}
\usepackage{algorithmic}
\usepackage{graphicx}
\usepackage{textcomp}
\usepackage{xcolor}
\usepackage{subcaption}

\usepackage{booktabs}

\begin{document}

\let\WriteBookmarks\relax
\def\floatpagepagefraction{1}
\def\textpagefraction{.001}

\shorttitle{Building a Cognitive Twin Using a Distributed Cognitive System and an Evolution Strategy}    

\shortauthors{Wandemberg Gibaut; Ricardo Gudwin}  

\title [mode = title]{Building a Cognitive Twin Using a Distributed Cognitive System and an Evolution Strategy}

\author[1]{Wandemberg Gibaut}

\cormark[1]

\fnmark[1]

\ead{wgibaut@dca.fee.unicamp.br}


\affiliation[1]{organization={University of Campinas (Unicamp)},
            city={Campinas},
          citysep={}, 
            state={São Paulo},
            country={Brazil},
            orcid={ https://orcid.org/0000-0001-7322-5399}}


\author[2]{Ricardo Gudwin}

\fnmark[2]


\ead[url]{https://faculty.dca.fee.unicamp.br/gudwin/}


\affiliation[2]{organization={University of Campinas (Unicamp)},
            city={Campinas},
            state={São Paulo},
            country={Brazil}}


\cortext[1]{Corresponding author}

\fntext[1]{}


\begin{abstract}
This work presents a technique to build interaction-based Cognitive Twins (a computational version of an external agent) using input-output training and an Evolution Strategy on top of a framework for distributed Cognitive Architectures. Here, we show that it's possible to orchestrate many simple physical and virtual devices to achieve good approximations of a person's interaction behavior by training the system in an end-to-end fashion and present performance metrics. The generated Cognitive Twin may later be used to automate tasks, generate more realistic human-like artificial agents or further investigate its behaviors.

\end{abstract}



\begin{keywords}
Cognitive Systems \sep  Artificial Intelligence \sep  Distributed Systems \sep Cognitive Twin \sep Internet of Things
\end{keywords}

\maketitle

\section{Introduction}
\label{sec:sec1}

This work proposes an approach that uses an evolutionary algorithm along traditional Machine Learning methods to build a digital, distributed cognitive agent capable of emulating the potential actions (input-output behavior) of a user while allowing further analysis and experimentation – at a certain level – of its internal structures. We focus on the usage of simple devices and the automation of this building process, rather than manually designing the agent. To accomplish this, we used DCT \citep{gibaut2020extending}, a tool to build distributed cognitive architectures across multiple physical and virtual devices. We argue that this brings some advantages, like true, massively parallel processing, low computing power device usage, and the exploration of different ways in which each part of the system might interact with others. In this introduction, we provide a primer on cognitive systems (subsection \ref{cognitivesystems}), system of systems (subsection \ref{systemofsystems}), Cyber-Physical Systems (\ref{cyberphysicalsystems}), and related work on cognitive twins (subsection \ref{relatedwork}). In section \ref{sec:sec2}, we describe DCT, providing an architectural overview of DCT, its codelets, and memory structure, introducing important terminologies to be used later in this article. In section \ref{sec:sec3} we describe our vision on how to evolve a cognitive twin, and in section \ref{sec:sec4} we describe our experiments. Finally, in section \ref{sec05} we discuss our results, provide some conclusions, and indicate future works related to this research.

\subsection{Cognitive Systems}
\label{cognitivesystems}

The research area of Cognitive Systems emerged in the early 90s, derived upon the research on Expert Systems, Production Systems, and Knowledge-Based Systems \citep{newell1961gps, simon1971human}. Investigating the origins of human knowledge, \cite{anderson1989theory} developed the ACT-R Cognitive Architecture \citep{anderson2004integrated,anderson2009can}. Approximately at the same time, based on the ideas presented by \cite{newell1989symbolic}, Laird released early versions of the SOAR cognitive architecture \citep{laird1996evolution,laird2012soar}. By the end of the 1990s, a large group of researchers involved in the \emph{Simulation of Adaptive Behavior} shaped the concept of Cognitive Architecture as an essential set of structures and processes necessary for the generation of a computational, cognitive model that could be used in several areas relating cognition and behavior. \cite{sloman2002architecture}, for example, discussed the ``architectural'' question of representing mental concepts, and \citet{sun2004desiderata} proposed a set of requirements needed to build a Cognitive Architecture. Next, \cite{sun2007importance} investigated issues and challenges in the development of such models. \cite{langley2009cognitive} analyzed several existing Cognitive Architectures, evaluating the overall progress in the research area. These studies indicated that the main advantage of characterizing a system as \emph{cognitive} is the possibility of obtaining a concrete \emph{framework} for cognitive modeling, in such a way that these models could be concretely tested. Such characterization might allow the definition of a set of structures, modules, and basic processes, creating a common language so that different researchers could explore their proposals for cognitive models.

Following this tradition, we understand a Cognitive System as a general-purpose control system inspired by scientific theories developed to explain the process of cognition in humans and animals. The idea of having an architecture is to decompose the cognitive phenomenon into a spectrum of capabilities, e.g. perception, attention, memory, reasoning, learning, behavior generation, and others. Furthermore, such architectures would also function as theoretical models of cognitive processes, allowing such theories to be tested and reused in different applications.

Cognitive architectures have been tested in several applications, from robot control to decision-making processes in intelligent agents, and substantially facilitate efforts to build intelligent artificial agents due to the specification of cognitive models inspired by human and animal cognition.

The study of cognitive architectures has become quite popular more recently, although several questions remain open. Nevertheless, over the last 20 years, several Cognitive Architectures have been proposed \citep{goertzel2014brief, kotseruba2016review}. In 2010, Samsonovich \citep{samsonovich2010toward} developed a comparative table presenting a systematic review of many known and documented cognitive architectures.

\subsubsection{Digital and Cognitive Twins}

In the past few years, due to the rise of new technologies and the challenges of a hyper-connected digital world, there is a growing interest in digital copies of physical entities, which are usually referred to as \emph{Digital Twins}. Although there is no consensus on the definition of a Digital Twin, a good definition is presented by \cite{lim2019state}, as a "high fidelity virtual replica of the physical asset with real-time two-way communication for simulation purposes and decision-aiding features for product service enhancement". 
There are many uses of a digital replica of an entity, and \cite{zhang2020towards} present a hierarchy that classifies Digital Twins from the simplest, descriptive type to the more complex ones. These are classified as Cognitive Digital Twins and can replicate, up to some level, human cognitive processes, acting with minimal or no human intervention. In this case, cognition means processes like self-awareness, dynamic knowledge acquisition, and management and decision-making capabilities in both digital and physical worlds, in an info-symbiotic manner.

Among other definitions, \cite{abburu2020cognitwin} present three levels of Digital Twins, being a Cognitive Twin (CT) defined as a hybrid, self-learning, and proactive system that optimizes its cognitive capabilities over time based on collected data and gained experience. It combines AI analytics techniques and expert knowledge to control the system's behavior, solving emerging problems.

In a more agent-centered vision, \cite{somers2020cognitive} understand a CT as a “digital reflection of the user, intended to make decisions and carry out tasks on the user's behalf”, to “highlight the key role that cognitive mechanisms play in modeling human decision-making in the IoT digital space”. There, they apply this concept as a way to model cognitive processes underlying the user’s decisions. In this same direction, \cite{du2020cognition} introduce a personal DT model of information-driven cognition (Cog-DT) as a “digital replica of a person’s cognitive process concerning information processing” including a VR platform and an adaptive UI.

As pointed out by \cite{abburu2020cognitive}, despite the existence of many definitions, there is an agreement that a Cognitive Twin is an extension of a Digital Twin with some cognitive capabilities, even though there is no consensus on what those capabilities might be. Each definition has its own set of capabilities, which better suit the area and application the definition came from.

\subsection{Systems of Systems}
\label{systemofsystems}

According to \cite{gudwin2016stateOfTheArt}, the term Systems of Systems (SoS, \emph{Systems of Systems}) began to appear in the 1990s and still today, continues to attract interest. Although being a widely discussed issue, the community has not yet converged to a common understanding of its meaning. According to  \citep{gorod2008system}, which tries to track the origin of the term, the first researcher to propose something related to the modern idea of an SoS was \cite{boulding1956sos}, who imagined SoS as a “gestalt” in theoretical construction creating a “spectrum of theories” greater than the sum of its parts. Later, \citet{jackson1984system} suggested the use of “SoS methodologies” as the interrelationship between different systems-based problem-solving methodologies in the field of operation research. Although the field has some pioneers, as shown above, it was only in 1989 that we find the first use of the term “system-of-systems” to describe an engineered technology system, in the Strategic Defense Initiative \citep{jacob1974logique, gpo1989restructuring, gorod2008system}.

From 1989 onward, modern research taking into account the term System of Systems was initiated. By that time, the concept of SoS was still very simple and poorly defined \citep{gorod2008system}. Among other definitions, Eisner \emph{et al.} \citep{eisner1991computer, eisner1993rcasse} defined an SoS as a set of systems (developed using a nominal systems engineering process) composed of interdependent systems combined to operate in a multifunctional solution, to achieve a common mission. \citet{shenhar1994new} defined an SoS as an ``array'' of systems, a large collection or network of systems working together to achieve a common goal. \citet{holland1995hidden}, in 1995, carried out studies on SoS from different perspectives. He described an SoS as an artificial complex adaptive system that, in continuous change, ``self-organizes'' with adaptive rules, to increase its complexity.
 In the same year, Maier proposed a characterization approach to differentiate a common system from a system of systems. According to him, an SoS must include ``operational interdependence of elements, managerial independence of elements, evolutionary development, emergent behavior, and geographic distribution''. Currently, Maier is one of the main contributors studying the field of SoS. In 1998, Maier proposed an SoS as a set of assembled components (which, in particular, are systems in themselves) exhibiting operational and managerial component interdependence \citep{maier1996architecting, maeir1998architecting}. \citet{kotov1997systems} and \citet{lukasik1998systems} then  addressed the issue of modeling an SoS. Further, \citet{boardman2006system} studied more than 40 SoS concepts, creating a common view on SoS. 

Following \citet{boardman2006system}, and also in consonance with others \citep{jamshidi2011system, gorod2014case, gorod2008system,dimario2006system,saunders2005system} there are 5 main characteristics shaping an SoS: Autonomy, Belonging, Connectivity, Diversity, and Emergence (or Evolution), usually referred as the ABCDE of SoS. Those characteristics may be described as follows:

\begin{itemize}
\item \textbf{Autonomy}: is the ability of an SoS component to make its own decisions independently of other systems present in its scope. Each system is free to make its own decisions. However, they cannot violate their performance issues and the SoS overall purpose. If a system violates the SoS overall purpose, it would be abandoned and, consequently, would make room for another system to join the SoS, instead of it.

\item \textbf{Belonging}: is the ability of systems to choose to belong to the SoS or not, based on their own needs and purpose. In other words, the system chooses to belong based on cost and benefit.

\item \textbf{Connectivity}: is the ability of an SoS component to maintain connections and exchange information with the other systems composing the SoS. There is a crucial problem here. How to ensure connectivity between systems, considering old legacy systems that do not change over time? How to add a new system to the platform? Researchers argue that it is necessary to define and build dynamic connectivity, creating interfaces and links to add and/or remove systems, in order to meet these needs. Also, an SoS might communicate with other systems autonomously and make connections in real-time.

\item \textbf{Diversity}: an SoS must be heterogeneous, i.e., its components might be of different kinds. If the SoS is diversified, it is more capable of working without becoming obsolete over time. Unlike common systems for which requirements should be defined even before they are designed, an SoS might (and should) update its requirements during runtime. With that, a heterogeneous SoS will remain stable for much longer, without having to be re-designed when new technologies become available. 

\item \textbf{Emergency}: is the ability to build new properties from developmental and evolutionary processes. With that, an unforeseen combination of system parts might result in a coalition that engenders a suitable interoperation of behaviors, which turns into an efficient solution to pursue the overall SoS goals. 
\end{itemize}

It is also important to note that an SoS cannot be developed based only on conventional systems engineering practices because an SoS has different characteristics, requiring a process of its own to deal with its unique particularities.

\subsection{Cyber-Physical Systems}
\label{cyberphysicalsystems}

During the 90s, different research communities were exploring the development of ``embedded systems''. From the integration between dedicated computer hardware and the automatic control of physical processes, a new technology started to emerge: \emph{Cyber-Physical Systems} (CPS).
According to \cite{lee2011introduction}, a CPS refers to embedded computers and networks monitoring and controlling physical processes, usually comprising feedback loops where physical processes affect computations and vice versa.

The term ``Cyber-Physical Systems'' was originally coined by Helen Gill, from the National Science Foundation, in the United States in 2006. The prefix \emph{cyber} in CPS comes from ``Cybernetics'', a transdisciplinary approach, defined by Norbert Wiener in 1948, for exploring regulatory systems, their structures, constraints, and possibilities. In other words: the study of the effects of feedback in systems theory to implement some sort of control on the system's variables.

There are many definitions of CPS in the literature, but the most common definition among contributors is that a CPS has to do with exposing parts of the urban environment to the Internet, where in some way, computational systems use environmental information collected by sensory devices to develop the control of environmental assets such that some goals are achieved, and some performance indexes are optimized. Also, CPS are defined as integrated systems that provide computation, networking, and physical processes. Another definition might be: a system where the physical environment and software components are tightly interrelated, each of them operating on different spatial and temporal scales, performing different behavior modalities, and interacting with each other. \citep{conti2012looking, sha2009cyber, horvath2012cyber, lee2009computing, nsf10515, khaitan2015design}.

According to \cite{miclea2011dependability} and \cite{khaitan2015design}, a CPS presents a set of fundamental characteristics:

\begin{description}
  \item[$\cdot$ Feedback Mechanism:] All physical components have a \emph{cyber} capability (feedback loop).
  \item[$\cdot$ Automation:] High levels of automation.
  \item[$\cdot$ Scalability:] Networking at multiple scales.
  \item[$\cdot$ Integrability:] Integration at multiple temporal and spatial scales.
  \item[$\cdot$ Reconfigurable:] Configuring and reconfiguring in real-time, dynamically.
\end{description}

The term \emph{cyber} can also be referred to the \emph{cyberspace}, i.e., the Internet, or the cyber world. Events in the real world need to reflect in the cyber world, and consequently, the cyber world needs to communicate with the physical world, such that decisions can be passed to actuators. This communication should be performed in real time and with high precision. A CPS needs to coordinate different systems (computing devices and distributed sensors/actuators) to act. In this case, sensors and actuators work like bridges between the physical and the cyber worlds \citep{khaitan2015design}.

In mid-2014, NIST established the CPS Public Working Group (CPS PWG) which aimed at gathering experts to reason about key aspects of CPS. This reflects the enormous interest in the area from a long-term perspective. Also, in Europe, there is a strong concern for continually raising awareness about Cyber-Physical Systems, since these are seen as one of the most promising technologies attracting European funding lately \citep{haakansson2015reasoning}.

Cyber-Physical Systems are inherently heterogeneous since they combine physical dynamics with computational processes. This heterogeneity exists even within the physical and cyber domains. The physical domain may be multi-physics, combining, for example,  chemical, and biological processes, mechanical motion control, and human operators. The cyber domain may combine programming languages, software component models, networking technologies, and concurrency mechanisms.
Cyber-Physical Systems are intrinsically concurrent. The cyber and the physical subsystems coexist in time, but even within these subsystems, concurrent processes exist. Models of concurrency in the physical world are very different from models of concurrency in software (e.g. arbitrary interleaving of sequences of atomic actions) and also very different from models of concurrency in networks (e.g. asynchronous, partially ordered, discrete actions or clock-driven time slots) \citep{lee2006cyber}. To conciliate these divergent models of concurrency, ensuring interoperability among components with different models of concurrency is one of the core issues to be handled in CPS.

\subsection{Related Work}
\label{relatedwork}

In \cite{abburu2020cognitive}, the authors apply a Cognitive Digital Twin to a real-world problem concerning predictive control in a steel-processing industry. Somers, Oltramari, and Lebiere \citep{somerscognitive, somers2020cognitive} use a Cognitive Architecture to build a Cognitive Twin as a personal assistant that learns user behavior from past data on a user's social network. Then, as a proof-of-concept case, it uses this knowledge to select attendees to a party. 

\cite{eirinakis2020enhancing} present the concept of Enhanced Cognitive Twin, in which they use Data Analytics and Machine Learning to provide them with a tool to solve environmental challenges, like decision-making capabilities, autonomous detection of anomalies and opportunities, and long-term optimization of and reasoning.

\cite{rozanec2020towards} use Cognitive Digital Twins in a manufacturing scenario to improve Key Performance Indicators (KPIs) by detecting possible anomalies and predicting their impact on production and by planning some aspects of production itself, like rescheduling existing plans.

In the Symbiotic Autonomous Systems White Paper II \citep{initiative2018symbiotic}, the authors put Cognitive Digital Twins as an almost essential asset to manage the knowledge gap, which would greatly impact future education.

 About using bio-inspired Evolutionary methods, \cite{zhang2020affective} applied a Classifier System in the context of Cognitive Robotics, building an emotional model for a robot. \cite{bellas2010multilevel} showed us a Cognitive Architecture that uses an evolutionary approach to provide an agent - hence a robot - with lifelong learning capabilities. That Architecture, the Multilevel Darwinist Brain, was also applied as a control method on two physical robots, Sony AIBO and Pioneer 2 \citep{bellas2006adaptive}.

To the best of our knowledge, we found no prior work on Cognitive Architectures and Evolution Strategy.

\section{The DCT}
\label{sec:sec2}

In this section, we will briefly explain our main tool. The DCT \citep{gibaut2020extending}, an acronym for Distributed Cognitive Toolkit, is a bare-bones toolkit to help the development of cognitive systems in a distributed, language-agnostic fashion. A cognitive agent created with DCT should be able to run across multiple physical (desktops, small computers like Raspberries, or microcontrollers like Arduino) or virtual devices (like Docker containers). It is a re-implementation of the ideas first seen in the CST main article \citep{Paraense201632}, as some features like being inherently single-device and being written in Java may be a shortcoming, sometimes. As expected, it also follows some theory lines like being Codelet-oriented, present in the Copycat architecture \citep{hofstadter1994copycat}. 

To better understand the DCT, one should first refer to the Cognitive Systems Toolkit (CST), the toolkit that came before. As its name suggests, CST is a toolkit for the development of cognitive architectures. Its purpose is to facilitate the creation of such systems the way the user wants, as long it respects its premises. In the core of CST are two basic entities that serve as building blocks: Codelets and Memories. A Codelet is a non-blocking, parallel process that runs continuously and represents a very specific piece of the cognition process of biological creatures. Likewise, a memory is a storage structure from which Codelets read and write information. A user may create any architecture that also follows a Codelet-oriented specification, based on already existing theories or new ones.

Many Cognitive Architectures like e.g. MECA \citep{gudwin2017multipurpose}, LIDA \citep{franklin2014lida}, and others may be seen as multi-agent systems. Following this paradigm, DCT conceives a distributed Cognitive System as a multi-agent system, where a standard protocol is used for the communication among the agents. Also, similar to its predecessor (CST), each agent in such a multi-agent system is built using Codelets and Memories.  The structure and functionality of a multi-agent system are fully compatible with distributed computing concepts and may be comfortably mapped to the Internet of Things: Sensory Codelets may be simply real-world sensor devices and Motor Codelets may be simply relays or actuators, while more complex Codelets may be embedded in microcontrollers or even software containers in the cloud.

\begin{figure*}[hbt]
\centering
	\includegraphics[width=2.0\columnwidth]{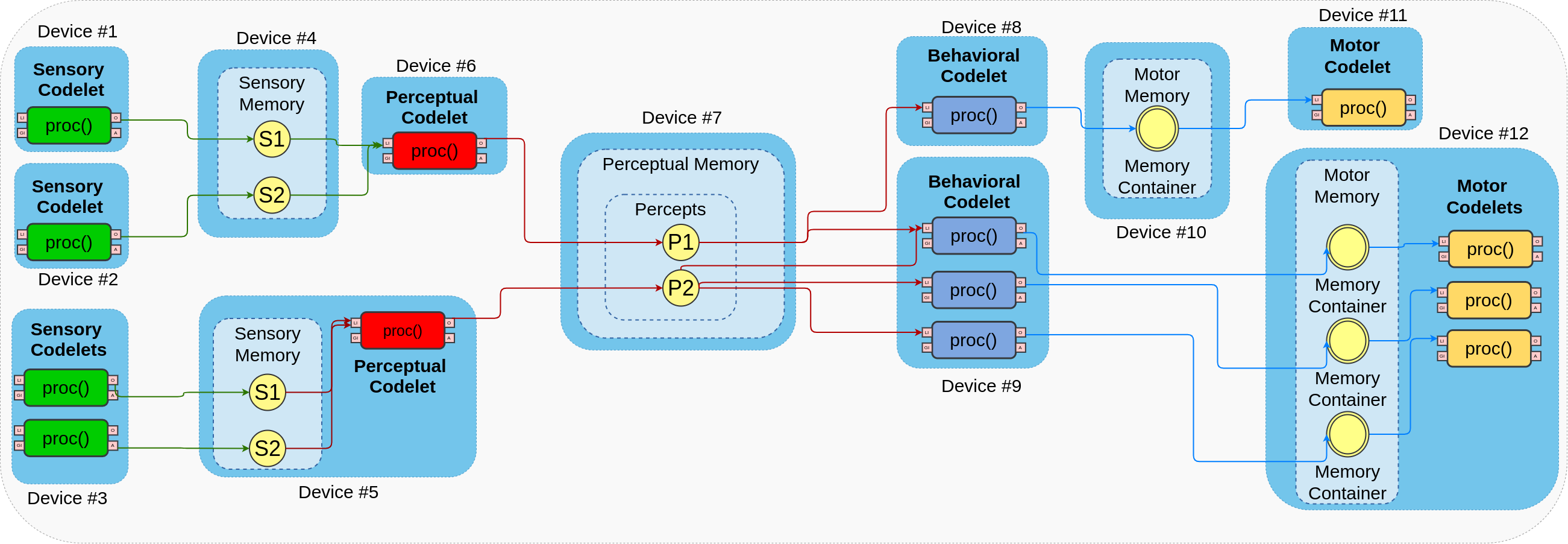}
\caption{Illustration of a multi-device, Codelet-oriented system as seen in \cite{gibaut2020extending}. Notice that, depending on how powerful is the device, it may run a single Codelet or multiple ones.}
\label{fig:multi_device}
\end{figure*}

Our first prototype of DCT - and current software based on it - is written in \emph{Shell script} and \emph{Python}, and could be deployed in containers (like Docker) or across different devices, including Raspberries and Arduinos. Figure \ref{fig:multi_device} illustrates the idea: if the input/output conditions are satisfied, there is no need to execute the whole system on a single computer. The project repository can be accessed in: \href{https://github.com/wandgibaut/dct}{https://github.com/wandgibaut/dct}.

\subsection{The DCT Architectural Overview}
\label{sec:dct_over}

From an architectural perspective, the DCT is composed of a set of \emph{Nodes} communicating to each other and integrating, as a whole, a functional system with cognitive capabilities. Theoretically, different subsets of the same collection of Nodes could even act as different systems. Here, the term \emph{Node} represents an entity (logical or physical), which works as a storage for groups of Codelets and/or Memories and is responsible for their operation and life cycles. Figure \ref{fig:multi_device} illustrates the idea of this non-homogeneity in device configurations. The subsection \ref{sec:dct_node} shows further details on this entity.

To communicate with each other, \emph{Nodes} follow a protocol regulating the interaction among them. Following CST specifications, Codelets only interact with Memories that is, a Codelet represents a block of computing unity, applying some process on data, but not holding it. This data storage is performed by Memories, which can be of different technologies. For this communication, DCT uses, canonically, \emph{json} formatted messages. This allows the use of a good range of technologies and simple sockets and databases like \emph{MongoDB} and \emph{Redis} are already supported by existing code. By following these directives, a user may use any language or technology suitable to a device in which the \emph{Node} is.

Formally, we can conceive a Cognitive System created with DCT in the following way:

\begin{definition}[A DCT Cognitive System]

Let $N$ be a set of \emph{Nodes}, where a \emph{Node} is an entity (logical or physical) that encapsulates one or more \emph{Codelets} and/or \emph{Memories} meant to be run under the supervision of a single \emph{Node Master} within an operational system.

To each \emph{Node}, there is an \emph{Interface} $I = \{MO, S\}$, where $MO$ is a subset of the \emph{Memories} implemented within a \emph{Node}, which will be accessed from other external \emph{Nodes} and $S$ is a Server that listens to a URI. This server $S$ should listen for requests and respond with \emph{json} formatted messages.

A Cognitive System created with DCT is defined by the interaction between the elements of $N$ following some Codelet-oriented Cognitive Theory, like MECA or LIDA.
\end{definition}

\subsection{The DCT Codelet structure}
\label{sec:dct_Codelet}

\begin{figure}[bt]
\centering
	\includegraphics[width=1.0\columnwidth]{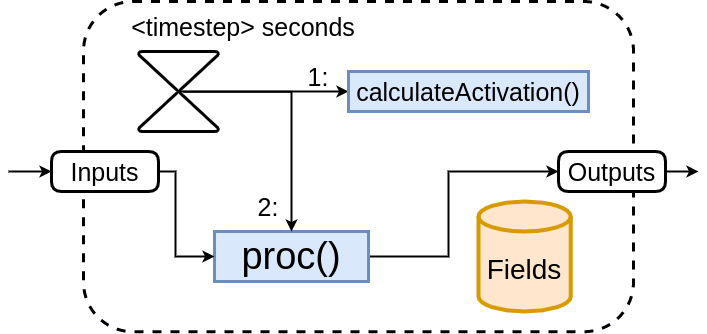}
\caption{The concept of a DCT Codelet.}
\label{fig:Codelet}
\end{figure}

A DCT \emph{Codelet} is composed of a callable program file in a user-specified language that follows some guidelines and some configuration files that can be used to dynamically change some properties, like which \emph{Memories} it can access. Figure \ref{fig:Codelet} illustrates this structure. It is valid to note that, since it was first implemented, some improvements have been made in how a \emph{Codelet} works. The files that characterize a \emph{Codelet} are:

\begin{itemize}
\item A \verb!Codelet! compiled program or script, which runs until be ordered (by the \emph{Node Master}) to stop
\item The \emph{Codelet} configuration file (\emph{fields.json}). This file contains some information regarding the \emph{Codelet} behavior, like its inputs, and should be possible to dynamically change it.
\end{itemize}

Also, if needed for a problem-specific reason, additional files may be used (a \emph{.ini} file, for example). In this work, \emph{Codelet} is implemented in \emph{Python} language.

Following the original CST implementation, the \emph{Codelet} program should have two main functions: a  \verb!calculateActivation! and a \verb!proc!. The first one is used to calculate the current relevance of the \emph{Codelet} itself and may be used in different ways, but mainly with a threshold value to decide if it should execute its main function or skip it. The second one performs the core functionality and is the most important function to be defined by the user. This function represents the procedural code that the entity will periodically call at each time step. Also, it should be non-blocking, meaning that it should be possible to run other processes alongside a \emph{Codelet}.

The configuration file (\emph{fields.json}) contains a structure that is analog to the \emph{Codelet} class in CST, defining many important parameters, e.g. the input, output, and broadcast ports, from where the Codelet can communicate to its Memory objects.

\subsection{The DCT Memory structure and default support}
\label{sec:dct_memory}

The other core structure, \emph{Memory}, is a generic term for the data structure that holds the information that \emph{Codelets} consume and/or process. Also, it contains some other meta-information, e.g. its name, URL, type, and an evaluation. As said before, this information is standardized as a \emph{json} structure. 

A \emph{Memory} should contain the following information:
 \begin{itemize}
     \item \emph{name}: String 
     \item \emph{IP/port}: String
     \item \emph{type}: String
     \item \emph{I}: String
     \item \emph{eval}: Double 
 \end{itemize}

\subsection{DCT Node}
\label{sec:dct_node}

In the DCT, a \emph{Node} is an abstraction for a physical or virtual device that contains an arbitrary number of \emph{Codelets} and/or \emph{Memories} and is supervised by a single \emph{Node Master}. This definition allows us to consider a computer to be a single \emph{Node} if all relevant entities run in the same environment, or to \textit{have} multiple \emph{Nodes} if each of them runs in a separated container with its \emph{Node Master}.
This \emph{Node Master} is responsible for starting, killing, adding, and removing Codelets and/or Memories, which are running through its supervision. Also, it should periodically check the health of its system, re-executing dead processes, and listen for external requests, like information requests or even requests to shut itself down.

Besides \emph{Codelets} and \emph{Memories}, a \emph{Node} should also implement an \emph{Interface} in which its internal entities may communicate with outside sources, e.g. a server with open sockets.

\section{An Evolutionary Cognitive Twin}
\label{sec:sec3}

Since we have already introduced some key concepts and the tool we're using, we can now discuss the proposed technique. Here, the main point is to build a Cognitive Twin using a vast amount of simple devices, orchestrated to work together as a single system, even if each device is a system of its own and may, theoretically, be a part of another system.

Here in this work, based on the definitions in section \ref{sec:sec1} and within the scope of what will be presented, the following definition will be presented:

\noindent\fbox{%
    \parbox{\columnwidth}{%
A \textit{Cognitive Twin} is a digital replica of the dynamics and cognitive - or just cognitive - processes of an intelligent physical system, usually aimed at a partial representation of a person. These cognitive processes refer to those identified in cognitive theories, such as perception, memory, behavior, adaptation, planning, learning, \textit{Reasoning} etc. The classification of an agent as \textit{Cognitive Twin} refers not only to the duplication of observable behavior of the virtual agent concerning the original but also to the possibility of in-depth investigation of the original individual through its copy.
    }%
}
\vspace{7mm}

The most fundamental idea here is the Codelet, already discussed in section \ref{sec:sec2}. We argue that sensors and actuators can be seen as Sensory and Motor Codelets, respectively. To make that consideration, we considered that both sensors and actuators are simple devices that do a very specific task, following the idea of Codelet, as seen in sections \ref{sec:sec1} and \ref{sec:sec2}. This consideration allows us to model our desired agent as a composition of simpler elements that interact with each other as needed.

Following that perspective, we postulate that the connections between sensors and actuators are given by some combination of elements that group and give them some sense, and elements that use that information to control actuators. This lets us use the concepts of Perceptual Codelets and Behavioral Codelets, respectively, largely used in other works that follow a Codelet-oriented Cognitive agency paradigm. Also, these premises fit in the concepts presented in section \ref{sec:sec1}: The System is composed of devices that are themselves systems connected through a network with an orchestration to build a Cognitive Agent that bridges virtual and physical domains.

But we have two main constraints: first, we sought to use simple, low computing power devices, and we do not have an infinite number of devices (with an infinite variety of input-output responses) to search for the best combination. We approach those constraints by having devices that can learn (or somehow adapt) and by making use of a heuristic that improves this search. It's also impractical to have those devices fully connected, as this communication overhead may degrade performance and/or be impossible if we deal with physical devices.

So, we propose to find the optimal configuration by having both explicit training on Codelets and an Evolution Strategy to find a suitable connection between Perceptual and Behavioral Codelets and between Behavioral and Motor Codelets. We'll discuss this in detail in the next subsections.


\subsection{Devices internal structure}
 First, we to define the internal structure of the devices we worked on. For the sake of simplicity, we used only virtual devices, as defined in section \ref{sec:sec2}, running Python Codelets on Docker. That allowed us to better manipulate some structures, like sending or requesting data from the master program to/from each Codelet and creating or destroying those virtual devices as we needed them. Even so, we kept the internal structures simple to draw a parallel with low-power devices. 
 
 As we mentioned before, we follow MECA Theory, which uses both \cite{Osman2004} theory of two, separated Cognition Systems working together and a \emph{Codelet}-oriented structure. Here we present a System 1 approach, which means that we'll be working only with four types of \emph{Codelets}: \emph{Sensory, Perceptual, Behavioral} and \emph{Motor Codelets}. We will briefly explain each of them.

 \subsubsection{Sensory Codelets}
 The simplest of them all, Sensory Codelets represent actual sensors, either physical or virtual. Like sensors, they are responsible for introducing the raw data into the system. For example, if we consider a human eye a sensor, the raw data is the light that enters the pupils. Here, it requests information about a specific attribute - temperature or luminosity, for example - of an environment. In this work, Sensory Codelets' internal structures will not be changed, simulating very simplistic devices, like a digital thermometer.
 
 \subsubsection{Perceptual Codelets}
 \label{sub:per}
 The subsequent structures in the information flow are Perceptual Codelets. These structures are responsible for aggregating the raw data that comes from Sensory Codelets in structures called Perceptions. In our human eye example, outputs of Perceptual Codelets would be \emph{depth}, \emph{objects}, relational properties (like distance from something), and so on. Notice that Perceptual Codelets, in a sense, represent how an agent experiences the world, as the information it could extract from data is heavily dependent on them. In this work, the internal structure of a Perceptual Codelet is represented by a Decision Tree Classifier, where the inputs are sensory data and the output is an integer value that represents a unique identifier (a token) of the input sensor's readings. Each Perceptual Codelet differs from another by the combination of its Sensor Codelets on the input. Those inputs are defined randomly with a uniform distribution to both quantity (a value between half and all the sensors) and which ones are picked up. 

 \subsubsection{Behavioral Codelets}
 Next, we have the Behavioral Codelets. The main purpose of these structures is to, based on previously structured information, activate one or more protocols to control what the agent should do, that is, to control one or more Motor Codelets. This Activation is usually encoded in a 0 to 1 \emph{float} value representing a Signal Strength, a way to measure how important that Behavior is to the current situation. Note that this so-called protocol may be anything from a simple heuristic to a whole Machine Learning method and the input of the Behavioral Codelet may include not only Perception but other information like those coming from a Motivational or even Emotional subsystem. Also, as a single Motor Codelet may have multiple Behavioral Codelets as input, those behaviors effectively compete to prevail and have their commands accepted.
   
   In this work, the Behavioral Codelets have also a Decision Tree Classifier as a method to decide what to send to its Motor Codelets based on Perception. 
   
 \subsubsection{Motor Codelets}
 The last basic structure here is the Motor Codelet. As mentioned before, this represents a direct parallel with an actuator, being physical or not. It simply responds to what was put as input and, through another Decision Tree, it sends a command to the corresponding entity in our virtual environment. This could be a direct association (''if this then that'') but, to make further usage of the code easier, we used a method that could accept more than one Behavioral without having to rewrite it completely.

\begin{figure}[ht]
\centering
	\includegraphics[width=1.0\columnwidth]{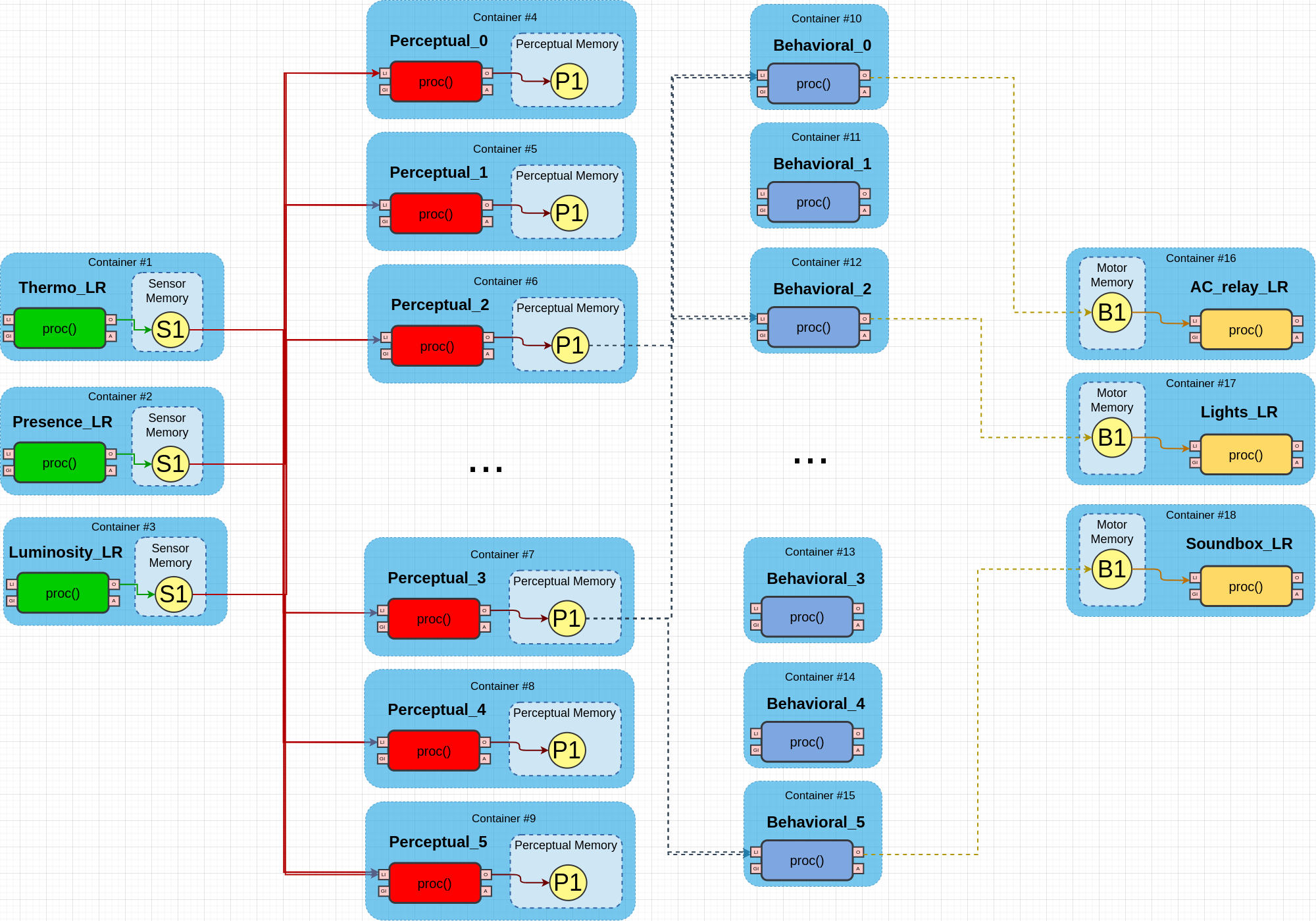}
\caption{Graphical representation of an agent structure and its internal connections}
\label{fig:agent_struct}
\end{figure}

 Figure \ref{fig:agent_struct} shows an overview of the structure of the topology of an agent. This concludes our overview of the main structures of our work. Next, we will detail about the Evolution Strategy.

\subsection{The optimization}

The general process of building the architecture for our Cognitive Twin involves determining the connection topology between the different types of Nodes and adjusting the internal functions of each Codelet to reproduce the overall behavior of a primordial agent. This is a two-step offline optimization process, the first being an optimization of the connection topology between Nodes and the second being a conventional training process of the Codelets' internal Machine Learning models in each Node. The connections between the Nodes will be defined through an evolutionary strategy and, given the connection configurations of each individual of a given generation, supervised training methods will be used to minimize the error between the expected and obtained outputs.
In this work, the internal structures of the Sensory Codelets are not changed, simulating very simplistic devices, such as a digital thermometer. 


\subsubsection{Evolution Strategy details}

In this part of the process, we want to, through an evolution process, define the best configuration of the connections between Perceptual and Behavioral Codelets and Behavioral and Motor Codelets. The connections between Sensorial and Perceptual Codelets are fixed and explained in subsection \ref{sub:per}.

\begin{figure}[hb]
\centering
	\includegraphics[width=1.0\columnwidth]{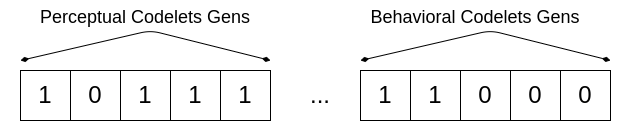}
\caption{Example Individual for our Evolution Strategy Process. The Individual is encoded as an array of binary values, each one representing if a certain Perceptual or Behavioral Codelet is to be considered as part of the agent.}
\label{fig:individual}
\end{figure}

To apply an Evolution Strategy, we need to do some definitions. First, we need to define our Individual encoding. Here, our Individual is a binary vector with the length of the number of total Perceptual Codelets plus the number of total Behavioral Codelets, where each index represents a specific Codelet. In that definition, a '1' represents that the corresponding Codelet is active on the Agent composition and a '0' means a non-connected Codelet. An example Individual is shown in figure \ref{fig:individual}.

Second, we need to define a mutation method. In this work, we adopted a simple 'bit-flip' probability for mutation, meaning that each individual has a probability \emph{mut\_p} to be mutated and each of its genes has a probability \emph{ind\_m} to change its state from '0' to '1' and vice versa. So, if a '0' becomes a '1', that means we should take the corresponding Codelet into account when mounting the agent topology.

As we adopted the output of Motor Codelets as exclusively binary, we can choose the fitness evaluation method as a Hamming Distance between the expected outputs and the actual ones. With this choice, the lower the Score, the better the individual fitness.

As a selection method, we choose to keep the best five individuals in the population for the next generation. Also, we choose the overall best one to be cloned. This procedure gives the possibility of recovering from local minima.


Also, we choose not to have any mating process. This choice was completely arbitrary, as we foresaw that it would not cause significant changes and required an additional process that could make each iteration longer.

\subsubsection{The training process}

To build our distributed agent correctly, we need a training process to ensure it accurately maps the system's inputs to the expected outputs. This training is done a) by changing its topology, choosing which Perceptual and Behavioral Codelets are composing the agent, and b) by fitting the data through all individual components consistently. 

\begin{figure}[ht]
\centering
	\includegraphics[width=0.7\columnwidth]{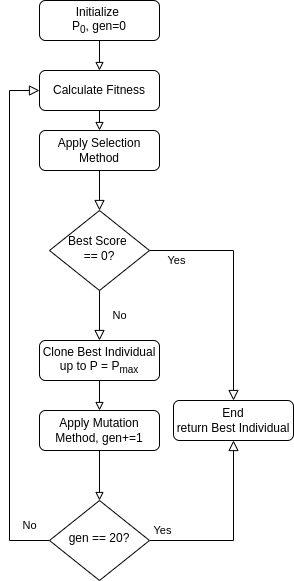}
\caption{Evolution Strategy process diagram. Here we have a high-level representation of each step of the mentioned heuristic.}
\label{fig:evol_strategy}
\end{figure}

The main component of the optimization process is very straightforward: it is a simple - yet efficient - Evolution Strategy to define the agent topology. This process is graphically represented in figure \ref{fig:evol_strategy}. But our \emph{evaluation} method requires more attention. It is in this part that we try to perform an input-output mapping.

First, we get our \emph{Individual} and reconfigure the Codelets connections properly, including cleaning all \emph{memories} and deciding which \emph{Behavioral Codelet} will feed each \emph{Motor Codelet}. We do that by writing a new \emph{fields.json} file and sending it to each \emph{Codelet}, and by forcing an empty value to the relevant \emph{Memories}, such as the \emph{motor-memories}, in each \emph{Motor Codelet}.

In the second part, we get our training inputs and send them to each relevant (the ones with a correspondent '1' in \emph{Individual} encoding) \emph{Perceptual Codelet} and send them a "train" signal, through a special \emph{Memory} on them with this sole purpose. This training process aims to create unique values - like tokens - that identify each observed combination of sensor readings. These readings refer solely to the sensors at the input of each \emph{Perceptual Codelet}. These tokens are integers corresponding to positions on an array with unique observations. For example, if ''[[0, 1, 0], [0, 1, 1], [0, 1, 0]]'' represents a set of training inputs, then ''[[0], [1], [0]]'' would be the outputs. Remember that each \emph{Perceptual Codelet} has its own sensor connections, so their responses differ one from another.

Now, the most sensitive part, we need to train \emph{Behavioral Codelets} properly considering the input-output response of each \emph{Perceptual Codelet} and respective \emph{Motor Codelet}. We do that in two steps: aggregating \emph{Perceptual Codelets} responses for each training input and mapping those responses to a \emph{Motor Codelet} input that would generate the desired output. These two sets (\emph{Perceptual} responses and \emph{Motor} input) represent the training we have on each \emph{Behavioral Codelet}.

The aggregation step is done by requesting the already trained model from \emph{Perceptual Codelets} and its respective input masks (representing which sensors feed them) and mounting a conjoined Perceptual output. This approach may also be useful for the data-sensitive task, as the system that collects/sends the information need not be the same that centralizes the Evolution Strategy.

Then, we need to get how each \emph{Motor Codelet} responds, either by getting a trained model or by getting an input-output set directly. We opt for the latter since \emph{Motor Codelets} represent actuators and, usually, they don't hold sensitive data. This part is problematic, as a \emph{Behavioral Codelet} may be assigned to two (or more) \emph{Motor Codelets} with fundamentally different behaviors. That could make it impossible to achieve correct model training. After collecting data, we send each input-output information to the respective \emph{Behavioral Codelet}.

Next, we are ready to evaluate the system's performance. One by one, we send the corresponding entry in the input test set to the environment server the system is sensing (we talk about this specificity in section \ref{sec:sec4}) and wait until the information propagates through a distributed agent, getting the system response after that (all \emph{Motor Codelets} outputs combined). This ''wait time'' is directly related to communication overheads and process concurrency if using a single computer.

Finally, we calculate the Hamming Distance between the expected values (test output) and the actual output. This will represent the \emph{Fitness} of the \emph{Individual}.

\section{Experiments}
\label{sec:sec4}

To test our model, we set up batch experiments that simulate interactions of an agent (that could be, for example, a human) with some devices in a smart home. Those devices are sensors and actuators. As sensors, we have, for example, presence detectors and thermometers, that generate the relevant data as the agent has its normal behaviors. As actuators, we some devices that would require direct interaction, like light switches, from which we get the agent preferences. Our goal was to map the readings of sensors and the activation of the actuator devices. We did find works that used interaction-centered data like \cite{engelmann2016interaction}, but none of them had a publicly available dataset.

Specifically, we simulated the house and its devices as a server that listens to a specific port for information and change requests. A sensor may request the current readings of its specific device, and an actuator may set a device ''on'' or ''off''.

As explained before, one of the main processes in the training of our Cognitive Twin is the Evolution Strategy, where we let a set of random choices of Perceptual and Behavioral Codelets be part of the agent. Each Perceptual differs from the other by the sensors it collects information: this may be any number between half of the available sensors and all of them, chosen randomly. Each Behavioral Codelet differs from others in the Motor Codelet (or Codelets) it feeds. 

Each individual is defined by an array of binary values representing the agent, so the mutation process is a "bit-flip" probability, and we choose not to use crossing-over methods. The selection method is "best five", meaning that we keep the best five individuals in each generation. The population at the beginning of each generation has twenty individuals and the initial population was randomly generated with each individual having a 20\% probability of '1' in each gene, which roughly reflects as a "mean participating number of Codelets" of 20\%.

The experiments were aimed to be relatively human-like, even if it is still synthetic. Based on the work of \cite{mendez2009simulating}, we generated data based on a transition probability matrix associated with interaction probability functions. So, our data is generated based on which room the agent would be in and with which devices it interacts. First, our base agent has some proprieties, like thermal comfort, that are the equivalent of a human having some preferences. These properties - and how their values are chosen - are:

\begin{itemize}
    \item \textbf{thermal\_comfort}: a uniformly random integer number in the interval [20, 30] that represents the temperature at which our base agent feels uncomfortable and turns AC on.
    \item \textbf{light\_comfort}: a uniformly random real number in the interval [0.4, 1] that represents the luminosity that our base agent feels uncomfortable and turns lights on.
    \item \textbf{voice\_prob}: a uniformly random real number in the interval [0, 1] that represents the probability that our base agent uses a voice-activated device.
    \item \textbf{coffee\_prob}: a uniformly random real number in the interval [0, 1] that represents the probability that our base agent uses the coffee machine, once in the room.
    \item \textbf{plant\_water\_th}: a uniformly random integer number in the interval [40, 100] that represents a threshold that a plant should be watered above.
\end{itemize}

\begin{figure}[bt]
\centering
	\includegraphics[width=1.0\columnwidth]{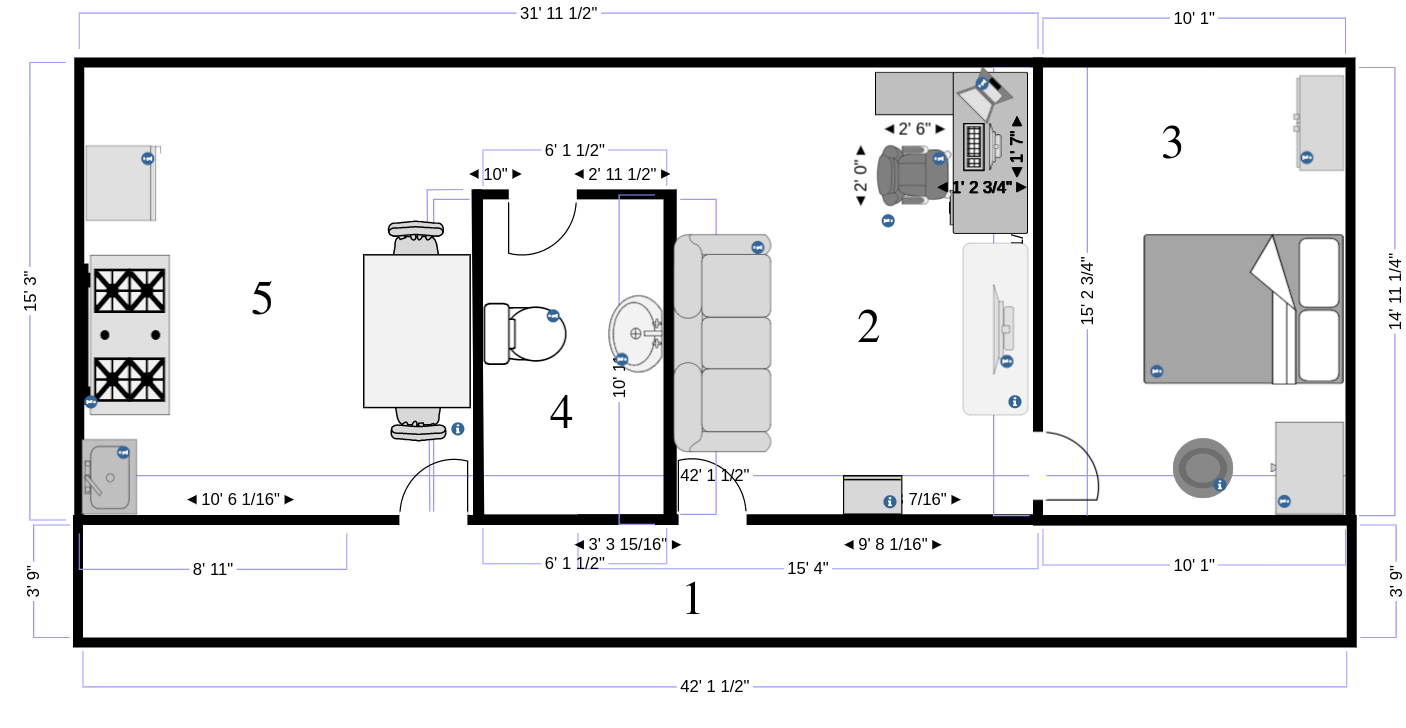}
\caption{House designed for the experiments. Notice this was used only to have a clear image when building an adjacency matrix}
\label{fig:house}
\end{figure}

Figure \ref{fig:house} illustrates the designed house. An agent could only move between consecutive rooms, as a human would do. This could be represented by an adjacency matrix. 
Considering that each non-null element of the adjacency matrix has an equal probability of being chosen, a transition matrix was built from it. This represents a ''room transition matrix'' and, in each room, the agent will have some interactions with room elements. These elements, being sensors and actuators, are shown in the list below alongside a brief explanation of how their states work.

\begin{itemize}
    \item \textbf{Existing sensors}
        \begin{itemize}
            \item  'presence': if the agent is on the room, it's '1'. Otherwise, is '0'
            \item  'temperature': a random integer number in [20, 30] interval
            \item  'luminosity': a random integer number in [0, 10] interval
            \item  'voice': a random binary choice with $p_1$ equal 'p\_voice'
            \item 'smoke\_detector': a random binary choice with probability of $p_1 = 0.1\%$ representing a fire occurrence
            \item 'shower\_temperature': a random integer number in [20, 30] interval
            \item 'humidity': a random integer number in [0, 10] interval
        \end{itemize}
    \item \textbf{Existing actuators}
        \begin{itemize}
            \item  'air\_conditioner': '1' if the 'temperature' sensor is lower than 'thermal\_comfort' and '0' otherwise 
            \item  'lights': '1' if the 'luminosity' sensor is lower than 'light\_comfort' and '0' otherwise   
            \item  'sound\_box': it's '1' if the 'voice' sensor was activated and '0' otherwise 
            \item 'coffee\_machine': a random binary choice with $p_1$ equal 'p\_coffee'
            \item 'anti\_fire': it's '1' if the ''smoke\_detector'' sensor was activated and '0' otherwise
            \item 'shower\_control': '1' if 'shower\_temperature' sensor is lower than 'thermal\_comfort' and '0' otherwise 
            \item 'plant\_watering': '1' if 'humidity' sensor is lower than 'plant\_water\_threshold' and '0' otherwise 
        \end{itemize}
\end{itemize}

After being in one specific state (one 'room'), all relevant sensors and actuator states change according to their behaviors. Lists of sensors and actuators each room has are shown below.
\begin{itemize}
    \item \textbf{living\_room}:
        \begin{itemize}
            \item sensors: 'presence', 'temperature', 'luminosity', 'voice'
            \item actuators: 'air\_conditioner', 'lights', 'sound\_box'
        \end{itemize}
    \item \textbf{bedroom}:
        \begin{itemize}
            \item sensors: 'presence', 'temperature', 'luminosity', 'voice'
            \item actuators: 'lights', 'sound\_box'
        \end{itemize}
    \item \textbf{kitchen}:
        \begin{itemize}
            \item sensors: 'presence', 'temperature', 'luminosity', 'smoke\_detector'
            \item actuators: 'lights', 'coffee\_machine', 'anti\_fire'
        \end{itemize}
    \item \textbf{bathroom}:
        \begin{itemize}
            \item sensors: 'presence', 'temperature', 'luminosity', 'shower\_temperature'
            \item actuators: 'lights', 'shower\_control'
        \end{itemize}
    \item \textbf{outside}:
        \begin{itemize}
            \item sensors: 'presence', 'luminosity', 'voice', 'humidity'
            \item actuators: 'lights', sound\_box', 'plant\_watering'
        \end{itemize}
\end{itemize}

For each run, we then took 400 samples based on a random walk using the transition matrix and the device interaction function shown above.
We ran a total number of 5000 experiments, in which we built slightly different base agents and different sample sets. This difference occurs in the randomly generated parameters reflecting ''preferences'', like \emph{thermal\_comfort}.

The experiments have a very specific main goal: to achieve an agent topology that presents the lowest (ideally zero) difference between expected output values and the actual ones. In the next section, we present the results and discuss them.

\section{Results and Discussion}
\label{sec05}

In this section, we present the results, discuss them, and make some conclusions about the experiments.

\begin{figure*}
        \centering
        \begin{subfigure}[b]{0.475\textwidth}
            \centering
            \includegraphics[width=0.8\textwidth]{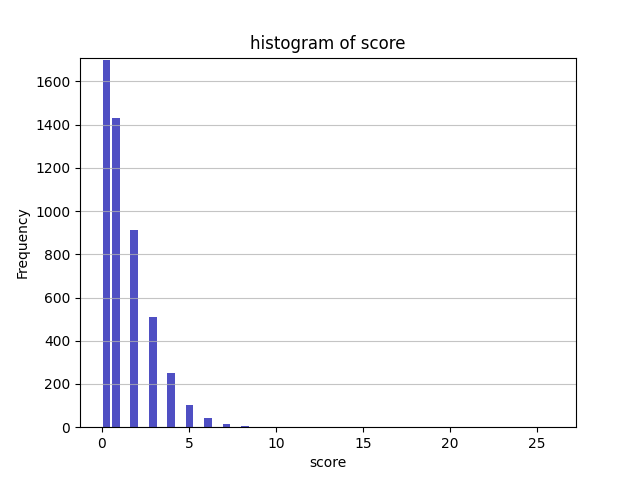}
            \caption[]%
            {{\small Histogram of \textit{score} achieved on experiment runs. Notice that the lesser, the better.}}    
            \label{fig:hist_score}
        \end{subfigure}
        \hfill
        \begin{subfigure}[b]{0.475\textwidth}  
            \centering 
            \includegraphics[width=0.8\textwidth]{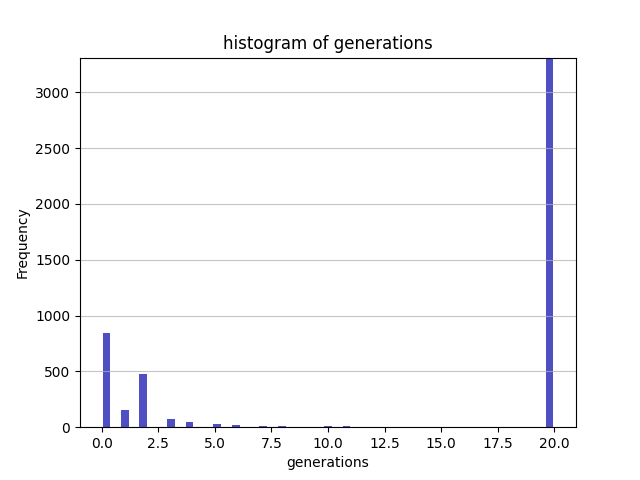}
            \caption[]%
            {{\small Histogram of \textit{generations} needed to achieve the lower score. Here, most runs needed the maximum number of generations}}    
            \label{fig:hist_gen}
        \end{subfigure}
        \caption[]
        {\small Histograms of the lowest score and generations needed to achieve that on each experiment run.} 
        \label{fig:hist_metrics}
\end{figure*}

With a slightly realistic scenario, the experiments present some interesting results. Figures \ref{fig:hist_score} and \ref{fig:hist_gen} show respectively histograms of (a) the final score after the full training process and (b) the number of generations the process took. Notice that most runs just stopped at 20 generations (maximum) and could not improve further, as Figure \ref{fig:hist_gen} suggests. Despite that, as can be seen in Fig. \ref{fig:hist_score}, more than 80\% of the runs ended with a score of 2 or less, meaning at most two wrong device activations on 260 interactions. 

\begin{figure*}
        \centering
        \begin{subfigure}[b]{0.475\textwidth}
            \centering
            \includegraphics[width=0.8\textwidth]{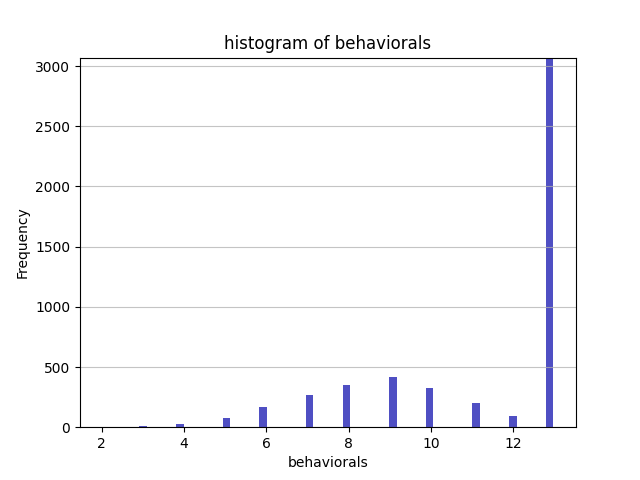}
            \caption[]%
            {{\small Histogram of the number of Behavioral Codelets needed to achieve the lowest score. Most runs needed 13, the same number of Motor Codelets (and actuators).}}    
            \label{fig:hist_beh}
        \end{subfigure}
        \hfill
        \begin{subfigure}[b]{0.475\textwidth}  
            \centering 
            \includegraphics[width=0.8\textwidth]{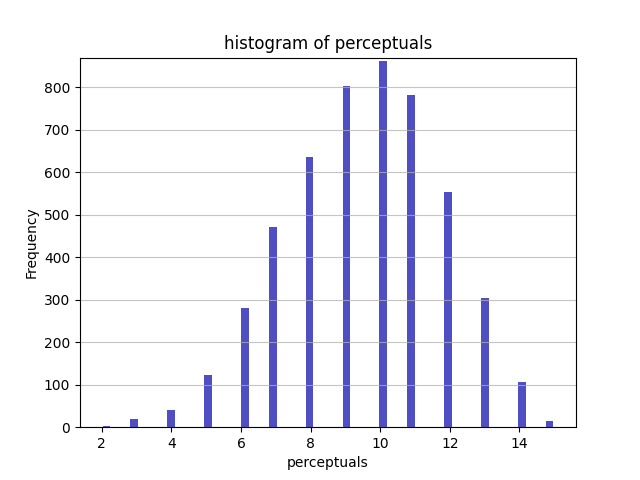}
            \caption[]%
            {{\small Histogram of the number of Perceptual Codelets needed to achieve the lowest score.}}    
            \label{fig:hist_per}
        \end{subfigure}
        \caption[]
        {\small Histogram of the number of ''internal'' Codelets needed to achieve the lowest score on each run.} 
        \label{fig:needed_codelets_2}
\end{figure*}

\begin{figure*}
        \centering
        \begin{subfigure}[b]{0.475\textwidth}
            \centering
            \includegraphics[width=0.8\textwidth]{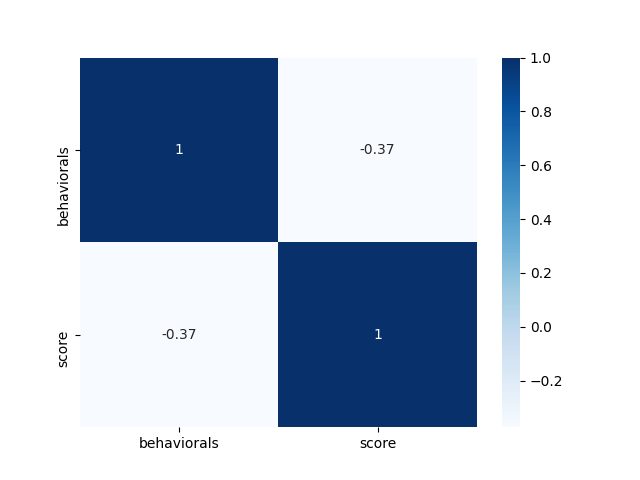}
            \caption[]%
            {{\small Correlation between number of Behavioral Codelets and Score}}    
            \label{fig:corr_behavior}
        \end{subfigure}
        \hfill
        \begin{subfigure}[b]{0.475\textwidth}  
            \centering 
            \includegraphics[width=0.8\textwidth]{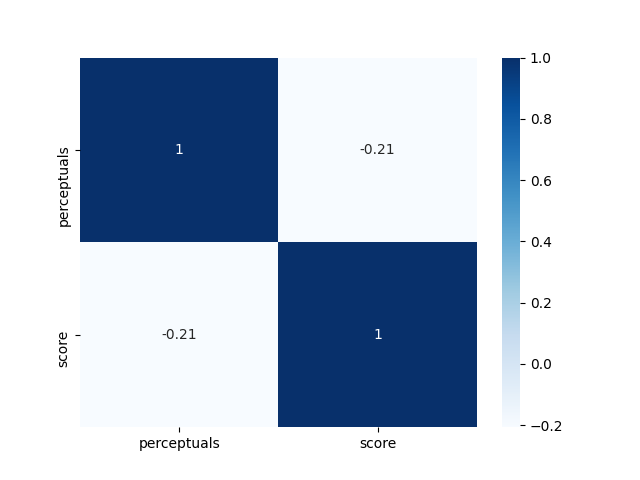}
            \caption[]%
            {{\small Correlation between number of Perceptual Codelets and Score}}    
            \label{fig:corr_per}
        \end{subfigure}
        \caption[]
        {\small Correlation between the number of ''internal'' Codelets and Score} 
        \label{fig:corr}
\end{figure*}

Figures \ref{fig:hist_beh} and \ref{fig:hist_per} reflect the number of Behavioral and Perceptual Codelets respectively to achieve the best result in each run. As we can see in Fig. \ref{fig:hist_beh}, most runs needed 13 Behavioral Codelets, one for each Motor Codelet/actuation device. Fig \ref{fig:corr_behavior} shows the correlation between the number of behavioral codelets and score. The system response is better (lower score) as more Behavioral Codelets are used.
The number of Perceptual Codelets, on the other hand, shows approximate normal distributions with a slight bias to the right, meaning that the embedding may vary and the output still be good. This bias is reflected in the slight negative correlation between the number of Perceptual Codelets and Score (smaller than Behavioral).

Table \ref{table:exp2} shows some statistics taken from the experiment. Notice that, while the individual number of Perceptual and Behavioral Codelets may go as low as 2, the combined ''Internal'' Codelets need a higher number to present satisfactory results.

\begin{table}[]
\centering
\caption{metrics on Experiments}
\label{table:exp2}
\begin{tabular}{@{}llllll@{}}
\toprule
              & mean    & median & std   & min & max \\ \midrule
score         & 1.438   & 1.0    & 1.894 & 0   & 26  \\
generations   & 13.6784 & 20.0   & 8.905 & 0   & 20  \\
n perceptuals & 9.5326  & 10.0   & 2.213 & 2   & 15  \\
n behaviorals & 11.2674 & 13.0   & 2.478 & 2   & 13  \\
n internals   & 20.8    & 22.0   & 3.998 & 7   & 28  \\ \bottomrule
\end{tabular}
\end{table}

\subsection{Conclusion and Future Works}

This paper has presented a pioneering approach to creating a Cognitive Twin by leveraging a distributed cognitive system in conjunction with an evolution strategy. Our work stands as a significant contribution to the field of cognitive computing by demonstrating the feasibility of orchestrating a multitude of simple physical and virtual devices to mimic a person's interaction behaviors. This achievement not only offers a practical application of distributed cognitive systems but also introduces a novel methodology for cognitive twin development, emphasizing the role of evolution strategies in optimizing system topology for more accurate behavior emulation.

In revisiting the themes introduced at the outset, our research seamlessly integrates the foundational principles of cognitive systems, Cyber-Physical Systems (CPS), and Systems of Systems (SoS) with contemporary advancements in artificial intelligence. By doing so, we have illustrated a comprehensive framework that not only addresses the complexities of human behavior simulation but also opens new avenues for automation, human-like agent creation, and in-depth behavioral analysis.

Comparatively, our approach distinguishes itself from established cognitive architectures such as ACT-R and SOAR, and the Standard Model of Mind, by emphasizing distributed processing and adaptability. While ACT-R and SOAR offer rich insights into cognitive processes through detailed psychological models, our model excels in harnessing distributed, interconnected devices to capture the multifaceted nature of human cognition. Similarly, the Standard Model of Mind provides a foundational framework for understanding cognitive functions. Yet, our work extends this understanding into the practical domain of CPS and distributed systems, offering a unique perspective on cognitive replication and interaction dynamics.

In conclusion, our research not only underscores the potential of distributed cognitive systems in creating sophisticated cognitive twins but also highlights the importance of evolutionary strategies in refining these systems. By drawing parallels and distinguishing our work from established cognitive architectures like ACT-R, SOAR, and the Standard Model of Mind, we contribute a novel perspective to the ongoing discourse on cognitive modeling and simulation.

Future work will focus on further refining the distributed cognitive system and exploring its integration with other AI paradigms and models. This research sets the stage for developing more sophisticated Cognitive Twins capable of performing complex tasks with minimal human intervention. By continuing to build on this foundation, future studies can enhance the fidelity and applicability of Cognitive Twins, making them tools in the field of cognitive computing.

\vspace{6pt}
\noindent\textbf{Declaration of Competing Interest}
The authors declare that they have no known competing financial
interests or personal relationships that could have appeared to influence
the work reported in this paper

\vspace{6pt}
\noindent\textbf{Acknowledgments:} This study was partially financed by the Coordenação de Aperfeiçoamento de Pessoal de Nível Superior - Brasil (CAPES) - Finance Code 001.

This project is part of the Hub for Artificial Intelligence and Cognitive Architectures (H.IAAC- Hub de Inteligência Artificial e Arquiteturas Cognitivas). We acknowledge the support of PPI-Softex/MCTI by grant 01245.013778/2020-21 through the Brazilian Federal Government

 \bibliographystyle{cas-model2-names} 
 \bibliography{main}

\end{document}